\documentclass[letterpaper, 10 pt, conference]{ieeeconf}  

\IEEEoverridecommandlockouts                              

\usepackage{graphicx} 
\usepackage{mathptmx} 
\usepackage{amsmath} 

\title{\LARGE \bf
Mechanically-Inflatable Bio-Inspired Locomotion for Robotic Pipeline Inspection
}

\author{Mostafa A. Atalla$^{1,2,^{*}}$, Fabian Trauzettel$^{1}$, Sebastiaan P. van Gelder$^{1}$, Paul Breedveld$^{1}$,\\ Michaël Wiertlewski$^{2}$ and Aimée Sakes$^{1}$
\thanks{This work was supported by a Delft University of Technology Cohesion grant. Fabian Trauzettel is part of the ATLAS project and as such, this work was supported by the ATLAS project. This project has received funding from the European Union’s Horizon 2020 research and innovation program under the  Marie Skłodowska-Curie grant agreement No 813782.}
\thanks{$^{1}$ Department of BioMechanical Engineering, Faculty of Mechanical Engineering, Delft University of Technology, 2628 CD Delft, The Netherlands}
\thanks{$^{2}$ Department of Cognitive Robotics, Faculty of Mechanical Engineering, Delft University of Technology, 2628 CD Delft, The Netherlands}
\thanks{$^{*}$corresponding author: {\tt\small m.a.a.atalla@tudelft.nl}}}

\begin{document}

\maketitle
\thispagestyle{empty}
\pagestyle{empty}

\begin{abstract}
Pipelines, vital for fluid transport, pose an important yet challenging inspection task, particularly in small, flexible biological systems, that robots have yet to master. In this study, we explored the development of an innovative robot inspired by the ovipositor of parasitic wasps to navigate and inspect pipelines. The robot features a flexible locomotion system that adapts to different tube sizes and shapes through a mechanical inflation technique. The flexible locomotion system employs a reciprocating motion, in which groups of three sliders extend and retract in a cyclic fashion. In a proof-of-principle experiment, the robot locomotion efficiency demonstrated positive linear correlation ($r=0.6434$) with the diameter ratio (ratio of robot diameter to tube diameter). The robot showcased a remarkable ability to traverse tubes of different sizes, shapes and payloads with an average of ($70\%$) locomotion efficiency across all testing conditions, at varying diameter ratios ($0.7\sim1.5$). Furthermore, the mechanical inflation mechanism displayed substantial load-carrying capacity, producing considerable holding force of ($13$~N), equivalent to carrying a payload of ($\approx 5.8$~Kg) inclusive the robot weight. This soft robotic system shows promise for inspection and navigation within tubular confined spaces, particularly in scenarios requiring adaptability to different tube shapes, sizes, and load-carrying capacities. The design of this system serves as a foundation for a new class of pipeline inspection robots that exhibit versatility across various pipeline environments, potentially including biological systems. 
\end{abstract}

\section{INTRODUCTION}
Pipelines are ubiquitous mean of fluid media transport, significantly impacting various aspects of our daily lives. They fulfill a multitude of roles, from the transportation of water, gases, and oils to the essential circulation of bodily fluids. These pipelines vary widely in size, ranging from sub-millimeter dimensions within biological systems to industrial counterparts measuring tens of centimeters. They also exhibit diverse curvatures, encompassing both straight and sharply curved configurations, and possess a wide array of structural variations, from rigid to soft. Their span extends over a broad range of distances, encompassing mere centimeters to extended lengths of hundreds of meters.  Despite the wide range of pipeline attributes, they all share a common necessity: regular inspections to ensure their proper functioning. For this purpose, several pipeline inspection robots have been developed targeting different applications.

Various inspection robots have been developed for pipeline navigation, encompassing wheel-type \cite{CHOI,kim}, walking-type \cite{Zagler}, crawler-type robots \cite{XU}, screw-driven type \cite{kakogawa} and other locomotion mechanisms \cite{Yeh,ismail}. These robots typically feature rigid structures, powered by electric motors and gear transmission systems, making them well-suited for inspecting pipelines with larger diameters. However, these designs exhibit certain limitations, such as a lack of flexibility that hinders their adaptability to pipelines of varying sizes. Downsizing these robots, particularly for millimeter-scale pipelines as found in biological systems, poses significant challenges. Furthermore, their intricate mechanisms necessitate complex assembly structures, which are not only maintenance-intensive but also prone to failure if debris enters their joints. The utilization of soft material can alleviate these drawbacks by rendering the robot completely soft.

\begin{figure}
    \centering
    \includegraphics{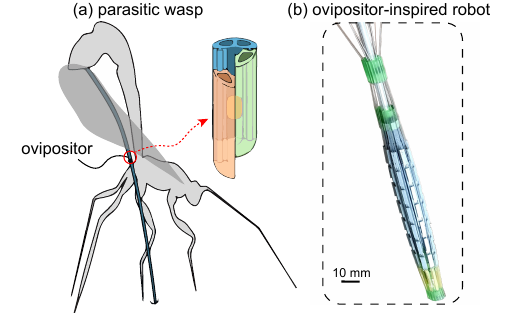}
    \caption{The working principle of the robot proposed herein is inspired by how a parasitic wasp uses its ovipositor to inject its eggs into a host medium. (a) illustration of the parasitic wasp, adapted from \cite{esther}. (b) prototype of the ovipositor-inspired robot proposed herein.}
    \label{fig:figure1}
\end{figure}

\begin{figure*}[!t]
    \centering
    \includegraphics{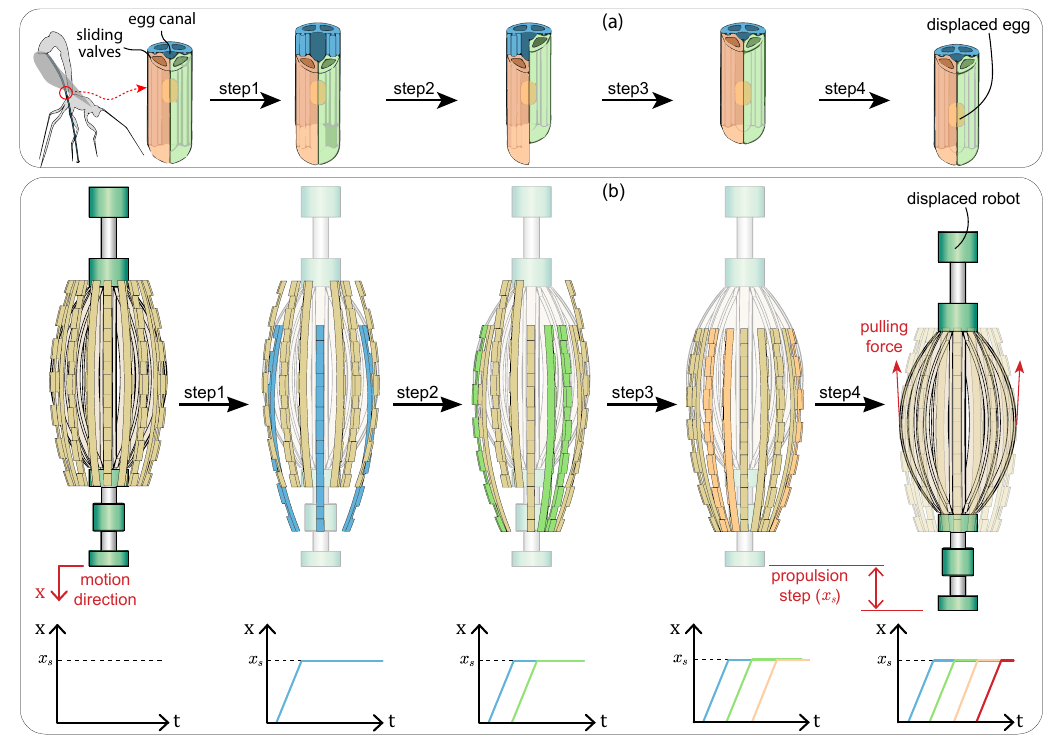}
    \caption{Illustration of the working principle of the wasp-inspired locomotion (a) the hypothesized motion sequence of the valves that the parasitic wasp uses to transport its eggs along its ovipositor, adapted from \cite{esther}; the valves retract sequentially, with one valve retracting and the other two remaining stationary, while the egg stays in place due to the higher friction of the stationary pair (step1-3). Once all the valves have been retracted, the wasp extends the entire valve assembly simultaneously, resulting in a displacement of the egg (step4). (b) the motion sequence of the ovipositor-inspired robot; a total of nine sliders are organized into three distinct groups, each consisting of three sliders. These groups operate in a coordinated, yet independent manner. The motion sequence initiates by advancing the first group of sliders forward, while the other two groups remain stationary. This pattern is then repeated with the other two groups until all groups have advanced into the forward position (step1-3). Subsequently, all three groups are pulled simultaneously, harnessing the cumulative friction force to propel the robot body forward, assuming the cumulative friction force surpasses the body inertial force (step4) (refer to the supplementary material video.1-2 for a demonstration of the motion sequence).}
    \label{fig:working_principle}
\end{figure*}

Soft robots have emerged as a promising technology for pipeline inspection. Their intrinsic flexibility enables them to conform to a variety of pipeline shapes and sizes, ensuring safe interaction with pipeline interiors, a particularly crucial feature in the context of biological pipelines. Most of the existing soft robotic designs for pipeline inspection rely on fluid for locomotion by harnessing a supply of air or hydraulic pressure to generate the motion, including inchworm-type \cite{shen,Yamamoto,adams}, earthworm-type \cite{Kundong,dew} and growing-type \cite{obregon} robots. Their relatively simple design offers a scalable alternative to rigid counterparts. Despite their advantages, fluidic-based designs share common drawbacks such as slower speeds and reduced durability. As robots traverse deeper into pipelines, fluidic response delays become more pronounced. Additionally, they are susceptible to rupture, especially in industrial pipelines where debris may be present. Hence, the quest for novel solutions that retain the merits of soft robots while addressing their drawbacks becomes evident.

Inspired by nature, we propose an ovipositor-inspired flexible locomotion system, Fig.\ref{fig:figure1}, that uses reciprocating flexible sliders to propel while adapting to tubes of different shapes and sizes through a mechanical inflation technique, pertaining the advantages of the soft structure while eliminating the downfalls of fluidic-based designs. In this study, we translated this principle into a fully functional prototype which we validated and tested experimentally. The remainder of this paper is organized as follows: in Section II, we provide an overview of the working principle, detailed design and prototype development of the proposed locomotion system. In Section III, we describe the experiments conducted to validate and characterize the developed prototype. Afterwards, we present and discuss the experimental results followed by a conclusion and outlook for the future in Section V.

\begin{figure*}
    \centering
    \includegraphics{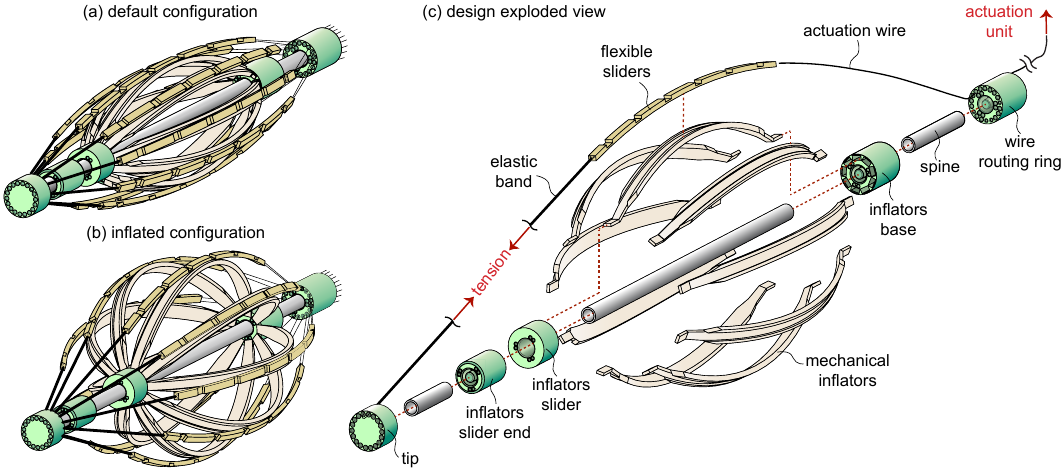}
\caption{Illustration of the robot locomotion system. (a) shows the locomotion system in its default (deflated) configuration. (b) shows the system in its inflated configuration. (c) shows the exploded view of the design featuring the main components of the system; the flexible sliders (a single slider is displayed for clarity), mechanical inflators and their associated actuation and structural support components (wires, sliders and rings).}
    \label{fig:design}
\end{figure*}

\section{Working Principle and Design}
An intriguing biological example of reliable mass transport within a tubular structure is evident in parasitic wasps. These parasitic insects are equipped with an ovipositor, an exceedingly thin and flexible tubular organ enabling them to penetrate living hosts for the purpose of depositing their eggs inside this host, as illustrated in Fig.\ref{fig:working_principle}(a). The transfer of eggs through the ovipositor is accomplished via an oscillatory motion of the ovipositor valves, which collectively constitute the tubular ovipositor \cite{esther, AUSTIN,cerkvenik,Meer}. The ovipositor typically comprises three valves that move independently: a dorsal valve and two ventral valves. Enveloped by an ovipositor sheath, these valves can slide longitudinally relative to each other while being secured radially through a tongue-and-groove connection \cite{AHMED}.

The state-of-the-art leading theories suggest that the friction between the valves and the eggs enables the transportation of the eggs along the ovipositor \cite{AUSTIN,AHMED}. One theory suggests that egg transportation and drilling occurs concurrently. In this theory, the motion sequence of the valves allows the ovipositor to penetrate deeper into the substrate while simultaneously transporting and depositing the eggs. Another theory posits that wasps transport their eggs via the ovipositor independent of penetration into the substrate. In this case, repeated small relative displacements of the ovipositor valves enable the transportation of the eggs, as depicted in Fig.\ref{fig:working_principle}(a). 

The sequence of egg transport, according to the second theory,  initiates with one of the valves retracting in the direction opposite to the intended transport direction, while the other two valves stay in place (Step 1 in Fig.\ref{fig:working_principle}(a)). Consequently, the egg stays in place because the collective friction between the two stationary valves and the egg surpasses the friction exerted by the single retracting valve, provided that each valve has the same contact area with the egg. Subsequently, the second valve retracts while the other valves stay in place  (step2 in Fig.\ref{fig:working_principle}(a)). Then, the last valve retracts while the rest stay in place  (step3 in Fig.\ref{fig:working_principle}(a)). Finally, all three valves translate simultaneously in the intended transport direction (step4 in Fig.\ref{fig:working_principle}(a)), where the egg advances in tandem with the valves due to friction. By this step, all valves will have returned to their initial positions, and the egg will have moved forward by one stroke length. By repeating this cycle, the egg can be transported along the entire length of the ovipositor until its deposition. 

Inspired by the egg transport mechanism of the wasp ovipositor, we designed a flexible self-propelling locomotion system, which comprises a wasp-inspired system of sliders to propel in tubular confined environments, integrated with a mechanical inflation system to adapt to different tube sizes and shapes. Fig.\ref{fig:design}(a) and (b) show the locomotion system in its default and inflated configurations respectively.

\subsection*{Self-Propelling Locomotion System}
In this design, the original arrangement of three sliding valves within the ovipositor case has been reconfigured into three groups of flexible sliders, as shown in Fig.\ref{fig:design}(a). Each group comprises three sliders, which operate collectively. This adaptation, from a single valve to a trio of sliders, enhances the robot capacity to conform to various lumen shapes. Each slider is connected to the tip through an elastic band at its distal end and is actuated through control wires attached to its proximal end, while resting on the mechanical inflators. The actuation wires are systematically organized and routed through a wire routing ring to constrain each wire to its corresponding slider. The structure of the locomotion system is held together using a rigid backbone (spine), which is a series of rigid tubes whose lumen can potentially allow for the deployment of inspection tools. 

The groups of sliders follow the same motion pattern as the ovipositor valves, as illustrated in Fig.\ref{fig:working_principle}(b). In the first step, one group of sliders advances forward, while the other two remain stationary. The greater static friction of the two stationary slider sets allows the moving group to progress without displacing the entire robot. This step is then repeated for the remaining two groups until all the sliders are in the forward position. Consequently, when all the slider groups are pulled simultaneously, the body moves forward by a single propulsion step ($x_{s}$), given that the static friction of all sliders surpasses the inertial force of the system. By repeating this sequence, the robot continues to move forward. This motion pattern is achieved through a cam and sliders mechanism in the actuation unit, with a cam profile that aligns with the motion sequence. The elastic bands, connecting the flexible sliders to the tip, play a crucial role in spring-loading the cam sliders, forcing them to follow the cam profile (refer to the supplementary material video.1-2 for a demonstration of the motion sequence). 

\subsection*{Mechanical Inflation System}
\begin{figure}[t!]
    \centering
\includegraphics{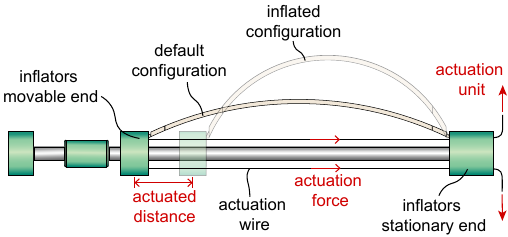}
\caption{Illustration of the mechanical inflation mechanism (with a single inflator for clarity). The mechanical inflators are mounted to a stationary end and a movable end. By sliding the movable end along the axis of the robot, the inflator undergoes an elastic deformation, resulting in radial mechanical inflation (refer to the supplementary material video.3 for a demonstration of the mechanical inflation mechanism).}
\label{fig:inflation_mechanism}
\end{figure}

The mechanical inflation mechanism comprises elastically deformable inflators affixed to a stationary end and a movable end. The movable end slides along the spine of the robot and is actuated via control wires, as depicted in Fig.\ref{fig:inflation_mechanism}. When those wires are pulled, the movable end moves closer to the stationary end. Consequently, the inflators undergo a radial elastic deformation, resulting in the mechanical inflation action shown in Fig.\ref{fig:inflation_mechanism}. Upon releasing of the wire tension force, the elastic energy stored within the inflators springs the movable end back, restoring its default deflated state (refer to the supplementary material video.3 for a demonstration of the mechanical inflation mechanism). 

To integrate the inflation mechanism into the locomotion system, we embedded the inflators beneath the flexible sliders which need to be in contact with the environment. To ensure the alignment of the inflators and the flexible sliders, we designed the inflators to have longitudinal channels which contrain the sliders laterally while allowing them to slide freely in the axial direction, as shown in Fig.\ref{fig:design}(c).

\subsection*{Prototype Development}
To validate the design, we implemented a low-cost proof-of-concept prototype. We optimized the design to allow for an extensive use of 3D printing technology for ease of fabrication. Consequently, all of the parts were 3D printed with the exception of the robot spine, for which we used off-the-shelf rigid aluminum tubes. The flexible sliders and mechanical inflators were 3D printed out of polylactic acid (PLA) filament material using a fused deposition modeling (FDM) 3D printer, Ultimaker3 (Ultimaker, Utrecht, The Netherlands). The rest of the components were printed using a resin-based stereolithography (SLA) 3D printer, Formlabs3B (Formlabs, Somerville, Massachusetts, U.S.A.). We employed off-the-shelf nylon wires to control the flexible sliders and mechanical inflators, and we used latex elastic bands to connect the robot's tip to the distal ends of the flexible sliders.

\section{Experiments}
To validate the concept of the mechanically-inflatable self-propelling locomotion mechanism and characterize its performance, we conducted two experiments. The first experiment serves as a proof-of-concept of the capability of the locomotion mechanism in traversing through tubes in different conditions. The second is to characterize the load-carrying capacity of the mechanical inflation mechanism.

\subsection*{Proof-of-Concept Experiment}
\begin{figure}[t!]
    \centering
    \includegraphics{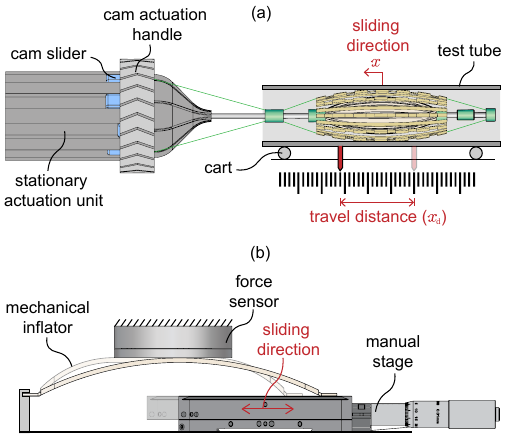}
    \caption{Illustration of the experimental setups. (a) shows the experimental setup used to characterize the efficiency of the locomotion system in different conditions (refer to the supplementary material video.4 for a demonstration of the proof-of-concept experiment). (b) shows the experimental setup used to characterize the holding force of the mechanical inflation system (refer to the supplementary material video.5 for a demonstration of the mechanical inflation characterization experiment).}
    \label{fig:setup}
\end{figure}

In this proof-of-concept experiment, we secured the robot prototype in place and positioned the test tube on a cart, as shown in Fig.\ref{fig:setup}(a). This arrangement allows the tube to move freely in the axial direction, while the robot itself remains stationary, minimizing operator influence on locomotion forces. To measure the travel distance of the tube, we affixed a pointer to the cart, which moved along a measuring tape, as depicted in Fig.\ref{fig:setup}(a). To resemble the weight of the robot while sliding, we created an offset between the axis of the robot and that of the tube, which results in a side pushing force equivalent to the weight of the robot (refer to the supplementary material video.4 for a demonstration of the proof-of-concept experiment). 

The experimental procedure started by positioning the robot entirely inside the tube and then recording the cart initial position. Subsequently, the robot was actuated to achieve one complete sliding cycle, after which the final position was recorded and the cart was reset to its initial position. This procedure was repeated three times for each testing condition. We conducted two sets of experiments using the same setup. First, we studied the performance of the robot across different combinations of tube sizes and robot radial inflations. Second, we studied the effect of loading, irregularity in the tube shape on the locomotion performance. 

To quantify the performance of the locomotion mechanism, we introduce two variables; the locomotion efficiency ($\eta_{l}$) and the diameter ratio ($\delta$). We define the locomotion efficiency as  the ratio between the distance traveled by the tube to the theoretical distance the tube could have traveled in case of ideal force transmission between the robot and the tube, expressed as follows:

\begin{equation}\label{eq:locEfficiency}
    \eta_{l} = \frac{|x_{final}-x_{initial}|}{x_{theoritical}}\times 100
\end{equation}

In the theoretical case, we assume a perfect slip between the flexible sliders and the tube when slider groups are individually advanced forward (step1-3 in Fig.\ref{fig:working_principle}(b)), and no slip when slider groups are all pulled back together (step4 in Fig.\ref{fig:working_principle}(b)). The diameter ratio ($\delta$) is the ratio between the diameter of the robot to the diameter of the tube, expressed as follows:

\begin{table}[b!]
    \centering
    \begin{tabular}{ccccccc}
    \hline
        actuated distance (mm)  & 0.3    & 0.9 & 1.5 & 2.1 & 2.7 & 3.3\\
        \hline
        robot diameter (mm) & 15 & 16.7 & 18.7 & 20.5 & 21.4 & 23.1\\
        \hline
    \end{tabular}
    \caption{}
    \label{table:radialexpansion}
\end{table}

\begin{equation}\label{slip_ratio}
    \delta = \frac{d_{robot}}{d_{tube}}
\end{equation}

To estimate the diameter of the robot, we measured the diameter in free condition (no contact condition) in relation to the actuated distance of the inflators' movable end depicted in Fig.\ref{fig:inflation_mechanism}. This distance-to-diameter mapping is summarized in Table.\ref{table:radialexpansion}.

\subsection*{Mechanical Inflation Characterization}
In this experiment, we characterized the holding force produced by the inflation mechanism. To do so, we affixed one end of a single slider to a manual linear stage and the other end to a fixture. We then measured the normal (holding) force generated by the mechanical inflation of the slider, actuated by the manual stage, using a force sensor touching the mechanical inflator, as shown in Fig.\ref{fig:setup}(b) (refer to the supplementary material video.5 for a demonstration of the mechanical inflation characterization experiment).

The experimental procedure started by fixing the slider to the manual stage such that the manual stage initial position is exactly where the mechanical inflation is about to occur. The force sensor is then brought in touch with the mechanical inflator such that the sensor is not loaded. We then actuate the inflator using the manual stage while recording both the normal force and the distance travelled. 

\section{Results and Discussion}
\begin{figure*}[!t]
    \centering
    \includegraphics{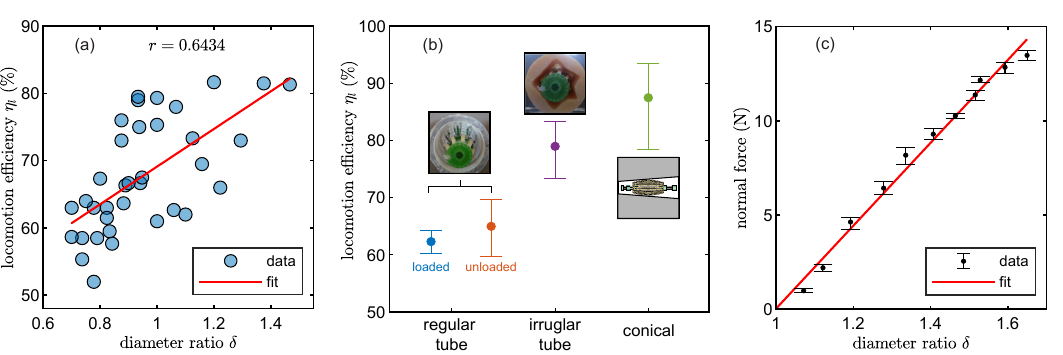}
    \caption{Experimental results. (a) shows the correlation between the locomotion efficiency ($\eta_{l}$) of the robot and the diameter ratio ($\delta$). (b) shows the locomotion efficiency ($\eta_{l}$) of the robot in tubes of different shapes and at different payloads. (c) shows the normal (holding) force of a single mechanical inflator as a function of the diameter ratio.}
    \label{fig:results}
\end{figure*}

The locomotion experiments revealed a relationship between the locomotion efficiency ($\eta_{l}$) and the diameter ratio ($\delta$) across different combinations of tube sizes and robot diameters.  Our findings revealed that the locomotion efficiency did not exhibit a direct correlation with either the robot size or the tube diameter alone. Instead, we observed a positive linear correlation ($r=0.6434$) between the locomotion efficiency and their ratio, the diameter ratio, as shown in Fig.\ref{fig:results}(a). This positive linear correlation is intuitive since as the diameter ratio increases, more sliders come into contact with the tube, leading to an expansion of the contact area and an increase in the transmitted pulling force, which consequently results in higher locomotion efficiency. Remarkably, the robot was able to propel at diameter ratios ($\delta<1$) with an efficiency of ($\eta_{l} \approx 60-70\%$), despite the robot size being smaller than the tube diameter, demonstrating its capability to propel without inflation. This result aligns with the wasp-inspired locomotion working principle, as only three of the nine sliders need to make contact, to propel. It is worth noting that at this range of diameter ratios ($\delta \leq 1$), the robot can only locomote in a horizontal tube configuration, and will not be able to propel through a vertical configuration. 

A noteworthy aspect of locomotion with a small diameter ratio ($\delta<1$) is the situation where only four or two sliders come into contact with the tube. In such instances, a specific occurrence takes place during the individual slider groups advancement steps (step1-3 in Fig.\ref{fig:working_principle}(b)): at one of these steps, an equal number of sliders move forward as sliders that remain stationary. This balance can lead to the elimination of friction asymmetry, potentially hindering the locomotion efficiency. Consequently, we anticipate that there might be certain combinations of tube sizes and robot diameters where efficiency is reduced when only two or four sliders are in contact. However, this particular condition has not been prominently observed in our experiments.

The following experiment aimed to assess the effect of cart weight on the locomotion efficiency. We conducted this experiment by loading the tube with a ($2$~Kg) weight, and then compared the results to the unloaded condition, all at a specific diameter ratio ($\delta=1$). The added weight introduced a higher friction requirement for tube movement. Unexpectedly, the experiment revealed that the load had no significant effect on locomotion efficiency. Specifically, the robot achieved an average efficiency of ($63\%$) under the loaded condition, as compared to ($65\%$) in the unloaded condition, as depicted in Fig.\ref{fig:results}(b). This outcome can be attributed to the robot slow accelerations and the relatively low friction of the cart used in the experiment.

It is important to note that this experiment solely investigated the impact of added inertia due to the weight. However, the weight of the robot itself can influence other aspects of the locomotion mechanism, such as the distribution of contact across the sliders. In cases where the robot weight increases, the load borne by the sliders at the bottom also increases, subsequently expanding their contact area compared to the other sliders. This effect can be mitigated by activating the mechanical inflation mechanism, which elevates the normal force across all sliders, resulting in a more equal distribution of friction forces.

Furthermore, we examined the capability of the robot to navigate through arbitrarily shaped tubes, simulating scenarios where tube obstructions might occur, such as due to the accumulation of contaminants. The robot demonstrated successful propulsion, achieving an average locomotion efficiency of ($79\%$) for the specific arbitrary shape illustrated in Fig.\ref{fig:results}(b). Our examination was not exhaustive and was only intended to offer a preliminary indication of the robot performance in such situations. Consequently, efficiency is expected to vary significantly depending on the specific tube shape. However, we anticipate that the inflation mechanism will mitigate potential performance limitations by conforming to the lumen shape. 

Contrary to our expectations, the experiment involving a conical tube yielded favorable results. We had anticipated that the conical shape would exert a counteractive force on the robot due to its shape, hindering the intended movement and reducing efficiency. Instead, the system exhibited the most favorable locomotion efficiencies among all tests. This unexpected outcome could be attributed to the increasing diameter ratio ($\delta=1.1\sim1.46$) along the length of the conical tube, which increases the contact area of the sliders as the robot propel inside the tube, resulting in an overall increase in the locomotion efficiency. This suggests that the system performs optimally when it has a larger contact area, which implies its potential effectiveness in a soft pipeline environment where the surroundings can conform to the robot shape. 

The mechanical inflation characterization experiment revealed a remarkable load carrying capacity for the inflation mechanism. Our findings indicated that a single inflator exhibited a stiffness of ($\approx 4.2$~N) per ($1$~mm) of actuation distance. With as little as 3mm of actuation, the robot could achieve a diameter ratio ($\delta=$1.6) and a holding force of ($\approx13$~N). In a hypothetical scenario where the robot is propelling in a vertical tube, the maximum load it can carry is determined by the cumulative frictional forces generated by its flexible sliders. These frictional forces can be estimated by multiplying the applied normal (holding) force with the static coefficient of friction ($\mu_{s}$) between the flexible sliders and the tubular environment. Assuming a coefficient of friction of ($\mu_{s}=0.5$), the robot has the capacity to transport approximately ($5.8$~Kg) of payload, which includes its own weight. This finding underscores the robot ability to excel in a vertical tube configuration, a scenario demanding robust load-carrying capacity to counteract the robot own weight and potential occlusion-induced forces. Furthermore, the ability to sustain high forces expands the range of tasks the robot can undertake, especially those involving force exertion. 

\addtolength{\textheight}{-10.5cm}   

\section{Conclusion and Future Work}
In this paper, we introduced a novel concept of an ovipositor wasp-inspired flexible locomotion system for robotic inspection in tubular confined spaces. Based on this concept, we designed a soft actuation mechanism that is able to adapt to different lumen sizes through a dedicated mechanical inflation mechanism, using the deformation of integrated elastic membranes. We implemented a proof-of-concept prototype which we validated experimentally. Experimental results showed that the locomotion mechanism is able to traverse through tubes of different diameter sizes, shapes and payloads with an average of ($70\%$) locomotion efficiency across all testing conditions at diameter ratios spanning from ($0.7$) to ($1.5$). Moreover, the mechanical inflation mechanism was able to produce ($\approx13$~N) of normal (holding) force which is capable of supporting ($\approx5.8$~Kg) of payload including the weight of the robot, in a vertical tube configuration, assuming a static coefficient of friction ($\mu_{s}=0.5$) between the flexible sliders and the tubular environment. 

In the future, we intend to advance the robot capabilities by creating a self-contained motorized version, featuring a flexible spine to enable steering and improved adaptation to varying pipeline curvatures. Subsequently, we aim to subject the robot to more realistic pipeline settings, evaluating its performance in real-world scenarios. Additionally, we plan to assess the robot locomotion efficiency in soft pipeline environments, with a specific focus on exploring its viability for medical applications.





\end{document}